\documentclass[journal]{IEEEtran}
\usepackage[utf8]{inputenc}
\usepackage{graphicx}
\usepackage{amsmath}
\usepackage{amsthm}
\usepackage{amssymb}
\usepackage{import}
\usepackage{color}
\usepackage{array}
\usepackage{diagbox}
\usepackage{subfig}
\usepackage{multicol}
\usepackage{multirow}
\usepackage{placeins}
\usepackage{hyperref}
\usepackage{balance}
\usepackage[normalem]{ ulem }
\usepackage{soul}
\usepackage{xurl} 
\usepackage{algorithm} 
\usepackage{algorithmic}  
\usepackage[algo2e]{algorithm2e} 
\usepackage{siunitx}
\usepackage[backend=bibtex,bibstyle=ieee,citestyle=numeric-comp]{biblatex}

\title{Deep Reinforcement Q-Learning for Intelligent Traffic Signal Control with Partial Detection}

\author{
\IEEEauthorblockN{Romain Ducrocq, Nadir Farhi} \\
\IEEEauthorblockA{COSYS-GRETTIA, Univ Gustave Eiffel, IFSTTAR, F-77454 Marne-la-Vallée, France}
}

\begin{document}
\maketitle
\begin{abstract}
Intelligent traffic signal controllers, applying DQN algorithms to traffic light policy optimization, efficiently reduce traffic congestion by adjusting traffic signals to real-time traffic.
Most propositions in the literature however consider that all vehicles at the intersection are detected, an unrealistic scenario. 
Recently, new wireless communication technologies have enabled cost-efficient detection of connected vehicles by infrastructures.
With only a small fraction of the total fleet currently equipped, methods able to perform under low detection rates are desirable.

In this paper, we propose a deep reinforcement Q-learning model to optimize traffic signal control at an isolated intersection, in a partially observable environment with connected vehicles. 
First, we present the novel DQN model within the RL framework. We introduce a new state representation for partially observable environments and a new reward function for traffic signal control, \\ and provide a network architecture and tuned hyper-parameters.
Second, we evaluate the performances of the model in numerical simulations on multiple scenarios, in two steps. At first in full detection against existing actuated controllers, then in partial detection with loss estimates for proportions of connected vehicles. \\
Finally, from the obtained results, we define thresholds for detection rates with acceptable and optimal performance levels. 

The source code implementation of the model is available at:

\textcolor{blue}{\textbf{\url{https://github.com/romainducrocq/DQN-ITSCwPD}}}
\end{abstract}


\section{Introduction}
\label{sec:intro}

\subsection{State of the art}

Traffic congestion poses serious economical and social problems; long travelling times, fuel consumption and air pollution; and inefficient traffic signals are significant underlying root causes to the issue. Fixed time traffic signals with predetermined timing have been commonly used, but become inadequate when facing dynamic and varying traffic demands. Thus, with ever growing urban areas and vehicle fleets, adaptive traffic signal control, able to respond to traffic flows in real time, is sought as a major stake of urbanization. \\

\subsubsection{DQN for traffic signal control}

Traffic signal control (TSC) has been addressed by reinforcement learning (RL) methods with Markov chains, dynamic programming, fuzzy logic \cite{alam2015design} and tabular Q-learning. However, the advent of deep Q-learning (DQN) in recent years has enabled to explore many novel TSC applications based on DQN algorithms. Initial studies with simpler models have proven DQN to be efficient for TSC, comparing trained multilayer perceptron models at isolated intersections with other algorithms from the literature \cite{stevens2016reinforcement,genders2019opensource}. Subsequent studies have build on these models to explore the benefits of complex neural network architectures, with stacked auto encoders, convolutional neural networks and recurrent neural networks \cite{li2016traffic,gao2017adaptative,liang2019deep,wei2018intellilight,vidali2019deep}. Latest work also successfully demonstrated a complete rainbow DQN implementation for TSC, with adaptations of the six DQN extensions \cite{alemzadeh2020adaptative}. Furthermore, a significant portion of the research in the field is dedicated to achieve decentralized coordination between traffic lights, with applications of multi agent reinforcement learning (MARL) over up to 1000 coordinated agents \cite{pol2016coordinated,wei2019presslight,wei2019colight,chen2020toward}. Yet, the main effort in the literature is put on defining state representations and reward functions. Indeed, the complexity of traffic environments leaves these definitions as unresolved, with many propositions. \\

\subsubsection{TSC with connected vehicles}

TSC responds to traffic demand based on real time measures of road traffic parameters. While the data are easily recovered in software simulations, real world implementations rely on expensive infrastructures, which exist only at a small fraction of intersections. These are mainly inductive loops under the roads, which allow for macroscopic representations of the traffic, or, in rare cases, radars or video cameras for microscopic descriptions. Most TSC propositions mentioned above require information that are thus difficult to obtain, and are for now mostly inappropriate. However, the rapid  development of IoT has created new technologies applicable for sensing vehicles, such as GPS localization systems, dedicated short ranged communications (DSRC), C-V2X, radio frequency identification, Bluetooth, ultra wide band, Zigbee, and apps (e.g. Google Maps) \cite{nguyen2020estimation,zhang2020using}. These communication devices are cost-effective, and do not require heavy infrastructure set ups. An increasing number of equipped connected vehicles (CV) are thus nowadays able to transmit their positions and speeds to infrastructures, allowing for microscopic partial state representations of intersections. Furthermore, latest research have shown that TSC with partial detection applied to CV penetration rates as low as 20\% can significantly improve traffic conditions for all vehicles \cite{zhang2019partially}. DQN applications for TSC with partial detection over connected vehicles have been little investigated, and are listed by surveys as a gap in the literature and a key future research.

\subsection{Paper contributions and organization}
In this paper, we apply a DQN algorithm to explore TSC policy optimization in partially observable traffic environments with limited detection on CVs, and the main contributions are:
\begin{enumerate}
  \item We review the traffic features used in DQN model definitions for TSC applications in the literature, sorted by agent actions, state representations and reward functions.
  \item We propose a new DQN model for TSC at single intersections with partial detection over connected vehicles. We introduce a new state representation for partially observable traffic, partial DTSE, and a new reward function for TSC, total squared delay. We provide tuned values for a DQN architecture and its hyper-parameters.
  \item We validate the model in full and partial detection with the 3DQN algorithm on three scenarios in the industry-standard SUMO simulator, by comparison against existing actuated TSC algorithms, Max Pressure and SOTL.
  \item We evaluate the performance loss in partial detection by proportions of CVs in the traffic, and estimate thresholds for the minimum acceptable and optimal detection rates.
\end{enumerate}

In section~\ref{sec:intro}, we introduce the subject with a state of the art of TSC with DQN and CVs. In section~\ref{sec:problem}, we lay the underlying assumptions, and formulate expressly the problem addressed. In section~\ref{sec:dqn}, we expose the background for Q-learning and DQN, detail our implementation of the 3DQN algorithm, and review the literature of DQN traffic features for TSC (contribution 1). In section~\ref{sec:model}, we propose the DQN model with the agent actions, state representation and reward function, and a DQN architecture with tuned hyper-parameters (contribution 2). In section~\ref{sec:methodology}, we present the evaluation methodology in SUMO with three simulation scenarios and two actuated comparison algorithms, Max Pressure and SOTL. In section~\ref{sec:results}, we validate the model with a comparative analysis of the results in full and partial detection (contribution 3), and conclude on the performances (contribution 4). In section~\ref{sec:conslusion}, we conclude on this paper and its perspectives.

\section{Problem statement}
\label{sec:problem}

\subsection{Assumptions}

\subsubsection{Road network} 
We consider a single 4-way intersection, where each leg has an incoming and an outgoing approach, and each approach has a set of incoming or outgoing lanes. Each incoming lane in an incoming approach has a set of connections to outgoing lanes in outgoing approaches. Each connection is controlled by one own traffic signal, and can be either open or closed. Incoming vehicles cross the intersection along open connections with right, left or through movements. \\

\subsubsection{Traffic phases} 
We consider a traffic phase as the complete sequence of a green phase with interval $T_g \geq T_{g,min}$, a change phase with fixed interval $T_y > 0$ and a clearance phase with fixed interval $T_r \geq 0$, with total phase interval $T_p = T_g + T_y + T_r$. The traffic program is the set of all possible phases, with a cycle being one complete rotation in a predetermined order through all the phases in the program. \\

\subsubsection{Connected vehicles} 
We consider CVs that can transmit their positions and speeds to traffic infrastructures. The mixed traffic is defined by a CV penetration rate $p_{cv}$, with a percentage $p_{cv}$ of CVs and a percentage $1-p_{cv}$ of non-CVs. We assume that CVs are perfectly observable, i.e. that their positions and speeds can be observed with perfect accuracy. This assumption is debatable; e.g. \cite{nguyen2020estimation} reports that current GPS localization systems are only precise enough to identify the approach in which a CV is situated, but not the exact lane.

\subsection{Problem formulation}
We formulate the problem addressed hereinafter as follows: \\
\textit{A DQN agent controls the traffic signals at an isolated intersection with the aim to minimize the total vehicle travel time, in a partially observable environment with limited detection over a fraction of the total mixed traffic from connected vehicles.}

\section{Deep reinforcement Q-learning}
\label{sec:dqn}

\subsection{Q-learning background}
The RL agent in stochastic environment $\varepsilon$ faces a decision making problem formalized as a Markov Decision Process (MDP), and defined by the tuple $<S,A,P,R>$, where $S$ is the continuous state space with state $s \in S$, $A$ is the discrete action space with action $a \in A$, $P$ is the transition dynamics matrix with $p=P(s'|s,a)$ the probability of transitioning to state $s'$ after taking action $a$ in state $s$, $R$ is the reward function with $r'=R(s,a,s')$ the reward of taking action $a$ in state $s$ and transitioning to state $s'$, and the transition is $(s,a,s',r')$.

The goal of learning in a MDP is to find a policy $\pi(s)$ mapping from states to actions, that maximizes the cumulative future reward $G_t = \sum_{k=t}^{t+T} \gamma^{(k-t)} \cdot r_{k+1} = r_{t+1} + \gamma \cdot G_{t+1}$, with $\gamma \in [0,1]$ a discount factor and a finite time horizon $T$.
    
The state-value function $V(s) = \mathbb{E}[G_{t} |s_t=s, \pi]$ estimates expected cumulative future rewards for successive states in the MDP with the Bellman recursion, and the action-value function $Q(s,a)= \mathbb{E}[r_{t+1} + \gamma \cdot V(s_{t+1}) |s_t=s, a_t=a, \pi]$ similarly estimates them for state-action pairs. For an optimal policy $\pi^*(s)$ at each step $t$, the Bellman optimality equation defines the relationship between optimal $V^*$ and $Q^*$ functions as $V^*(s) = \max_a Q^*(s,a)$, and thus the optimal action-value
\[ Q^*(s,a) = \mathbb{E}[r_{t+1} + \gamma \cdot \max_{a'}Q^*(s_{t+1},a')|s_t=s,a_t=a,\pi^*] \]

Q-learning algorithms approximate the $Q$-function by many iterative updates, with a function approximator $\Theta$ such that $Q(s,a,\Theta) \rightarrow Q^*(s,a)$ when $t \rightarrow \infty$. The agent then learns an optimal policy over time $\pi(s) = arg \max_a Q(s,a,\Theta)$ \cite{sutton2018reinforcement}.

\subsection{3DQN implementation}

Deep Q-learning algorithms approximate the $Q$-function with a neural network, the deep $Q$-network (DQN) \cite{hessel2017rainbow,mnih2013playing,mnih2015human} $Q$ with weights $\Theta$, mapping from one continuous state $s \in S$ in an $|s|$-dimensional input layer to all the discrete $Q$-values $Q(s,a,\Theta) \:\: \forall a \in A$ in an $|A|$-dimensional output layer. Here, we implemented the dueling double DQN (3DQN) algorithm: \\

\subsubsection{The dueling architecture} \cite{wang2016dueling} decouples the state-value function $V(s)$ and the action-advantage function $\widehat{A}(s,a) = Q(s,a) - V(s)$, into a 1-dimensional value stream $V(s,\theta,\kappa)$ and an $|A|$-dimensional advantage stream $\widehat{A}(s,a,\theta,\iota) \:\: \forall a \in A$, with weights $(\theta,\kappa,\iota) = \Theta$. The streams are then recombined in a special aggregation layer to produce the $Q$-function with
\[ Q(s,a,\Theta) = V(s,\theta,\kappa) + (\widehat{A}(s,a,\theta,\iota) - \frac{1}{|A|}\sum_{a'} \widehat{A}(s,a',\theta,\iota) ) \]

\subsubsection{The temporal difference} (TD) errors $\delta$ are computed from double TD targets \cite{hasselt2015deep} by forward propagation through the $Q$-network $Q$ and a target $Q$-network $\widehat{Q}$ with weights $\Theta^-$
\[ \delta = r' + \gamma \cdot \widehat{Q}(s', arg \max_{a'} Q(s',a',\Theta), \Theta^-) - Q(s,a,\Theta)\]
and the target $Q$-network $\widehat{Q}$ is updated by Polyak updates \cite{lillicrap2019continuous}
\[ \Theta^- = (1 - \tau) \cdot \Theta^- + \tau \cdot \Theta \text{, with } \tau \ll 1\]

\subsubsection{The replay memory} $D$ stores the $N$ last transitions, and the online $Q$-network $Q$ is updated by back propagation on a batch $U(D)$ of $M$ past transitions $(s,a,s',r')_m$ drawn uniformly at random from $D$, with Huber Loss over TD errors
\[ 
L = \frac{1}{2M}\sum_{m=1}^M
\begin{cases}
  (\delta_m)^2 & \text{if } |\delta_m| < 1 \\
  2 \cdot |\delta_m| - 1 & \text{else}
\end{cases}
\]
and with Adam optimization \cite{kingma2017adam} on the derived gradient $\Delta$.

\begin{algorithm}[H]
\caption*{Dueling Double DQN (3DQN) algorithm}
\begin{algorithmic}
    \STATE Initialize step $t = 0$;
    \STATE Initialize online dueling $Q$-network $Q$ with random weights $\Theta_0$ and target dueling $Q$-network $\widehat{Q}$ with weights $\Theta^-_0 = \Theta_0$; \\
    \STATE Initialize replay memory buffer $D$ to capacity $N$
    \STATE with $Nmin$ random transitions $(s,\text{rand }a \in A,s',r')$;
    \FOR{episode e = 1:E}
        \STATE Initialize sequence, observe initial state $s_t=\phi(x_t)$;
        \WHILE{$s_t$ not terminal}
            \STATE With probability $\epsilon$ select random action $a_t \in A$
            \STATE otherwise select action $a_t = arg\max_{a}Q(s_t,a,\Theta_t)$;
            \STATE Execute $a_t$ in emulator $\varepsilon$ and
            \STATE observe reward $r_{t+1}$ and next state $s_{t+1}=\phi(x_{t+1})$;
            \STATE Store transition $(s_t,a_t,s_{t+1},r_{t+1})$ in $D$;
            \STATE Sample batch $U(D)$ of $M$ transitions $(s,a,s',r')_m$;
            \FOR{m = 1:M}
                \STATE Set double TD error $\delta_m$;
            \ENDFOR
            \STATE Update online $Q$-network $Q$ with Adam optimization \\
            on Huber loss $L_t$ w.r.t. weights $\Theta_t$ and gradient $\Delta_t$;
            \STATE Soft update target $Q$-network $\widehat{Q}$ with Polyak update;
            \STATE Decay $\epsilon$ with exponential decay, increment step $t$;
        \ENDWHILE
    \ENDFOR
\end{algorithmic}
\end{algorithm}

\subsection{Review of DQN for TSC}

The main effort in the literature of TSC with DQN is placed on the definition of efficient state representations and reward functions. However, RL models are time consuming to tune, and the complexity of TSC environments makes evaluation difficult. Moreover, the issue of reproducibility in RL prevents rigorous comparison of TSC models, and there are no widely accepted representations. We compile here a non-comprehensive list of propositions found in the literature: \\

\subsubsection{Agent actions} 
Two types exists; either (1) the TSC agent acyclically selects the next phase from the set of possible phases with fixed green interval duration \cite{stevens2016reinforcement,li2016traffic,gao2017adaptative,vidali2019deep,alemzadeh2020adaptative,pol2016coordinated,wei2019presslight,wei2019colight,chen2020toward,zhang2020using,zhang2019partially,genders2018evaluating,touhbi2017adaptative}, or (2) the TSC agent selects the duration of the next green interval for the upcoming phase in a cycle \cite{genders2019opensource,liang2019deep,wei2018intellilight}.
\\

\subsubsection{State representation}
The observations can either be macroscopic, with estimations of global parameters over traffic flows, or microscopic, with raw data for individual vehicles. While earlier work preferred low-dimensional macroscopic state representations, latest work observe individual vehicles for high-dimensional microscopic state representations. Indeed, a significant gain in performance has been measured for microscopic observations over macroscopic ones \cite{genders2018evaluating}. Either way, state representations have been usually found to be aggregations of multiple traffic features; among which current phase \cite{genders2019opensource,gao2017adaptative,wei2018intellilight,pol2016coordinated,wei2019presslight,wei2019colight,chen2020toward,zhang2020using,zhang2019partially}, number of vehicles \cite{genders2019opensource,wei2018intellilight,alemzadeh2020adaptative,wei2019colight,zhang2020using,zhang2019partially,genders2018evaluating}, positions of vehicles \cite{gao2017adaptative,liang2019deep,wei2018intellilight,vidali2019deep,pol2016coordinated,genders2018evaluating}, queue lengths \cite{genders2019opensource,li2016traffic,wei2018intellilight,genders2018evaluating,touhbi2017adaptative}, speeds of vehicles \cite{stevens2016reinforcement,gao2017adaptative,liang2019deep,genders2018evaluating}, history of past phases \cite{stevens2016reinforcement,alemzadeh2020adaptative,genders2018evaluating}, elapsed time in the current phase \cite{zhang2020using,zhang2019partially}, green, change and clearance intervals duration \cite{zhang2020using,zhang2019partially}, distances to nearest vehicles \cite{zhang2020using,zhang2019partially}, pressures \cite{wei2019presslight,chen2020toward}, number of waiting vehicles \cite{stevens2016reinforcement}, waiting times of vehicles \cite{wei2018intellilight}, and upcoming phase in a fixed cycle \cite{wei2018intellilight}. \\

\subsubsection{Reward function}
The reward in TSC, overall, aims at reducing total lost travel time caused by traffic signals. As such criterion can not be assessed directly, rewards are mostly defined as weighted combinations of traffic features, which coefficients are set empirically. The lack of solid theoretical justifications is a major challenge for real world deployment, as RL rewards are difficult to transcribe mathematically without distorting the intended goal, especially in complex systems. Here, the reward functions act in effect as punishments, with agents reducing evaluation metrics derived from traffic features, maximizing negative values; among which delays of vehicles \cite{genders2019opensource,wei2018intellilight,pol2016coordinated,zhang2020using,zhang2019partially,genders2018evaluating,touhbi2017adaptative}, waiting times of vehicles \cite{gao2017adaptative,liang2019deep,wei2018intellilight,alemzadeh2020adaptative,pol2016coordinated}, queue lengths \cite{li2016traffic,wei2018intellilight,wei2019colight,touhbi2017adaptative}, intersection throughput \cite{stevens2016reinforcement,wei2018intellilight,alemzadeh2020adaptative,touhbi2017adaptative}, phase changes \cite{wei2018intellilight,alemzadeh2020adaptative,pol2016coordinated}, pressures \cite{wei2019presslight,chen2020toward}, accumulated waiting times of vehicles \cite{vidali2019deep}, number of waiting vehicles \cite{alemzadeh2020adaptative}, travel times of vehicles \cite{wei2018intellilight}, emergency stops  of vehicles \cite{pol2016coordinated}, and teleports of vehicles in SUMO \cite{pol2016coordinated}.

\section{DQN model for TSC}
\label{sec:model}

\subsection{Agent actions}

\subsubsection{Actions}
The agent selects the next phase from the set of possible phases for the next $T_{g,min}$ time of green interval. If the selected phase is the ongoing phase, the current green interval $T_g$ is extended by $T_{g,min}$. Else, the phase is changed to the selection, with an intermediate $T_y+T_r$ time through change and clearance intervals, and an initial $T_{g,min}$ time in the green interval of the new phase. Thus, while the action is to select the next phase, the agent also decides the duration of the ongoing phase by consecutively selecting the same phase.
\\

\subsubsection{Action space}
The action space is defined by the number of possible phases in the traffic program. At a four-way intersection, there is a total of eight valid paired signal phases for non-conflicting movement signals \cite{wei2020survey}. From this, and as phases aim to separate conflicting connections so that concurrent movements have minimum conflicts, the set of possible phases has either two or four phases. 2-phases programs have two permissive green intervals, one for each axis, with permissive turn left, through and turn right movements. The action space is $A=\{(n\rightarrow esw,s\rightarrow wne),(e\rightarrow swn,w\rightarrow nes)\}$, with size $|A|=2$. 4-phases programs have four protected green intervals, two for each axis, with protected turn left movements only, or through and turn right movements. The action space is $A=\{(n\rightarrow sw,$ $s\rightarrow ne),(n\rightarrow e,s\rightarrow w),(e\rightarrow wn,w\rightarrow es),(e\rightarrow s,w\rightarrow n)\}$, with size $|A|=4$.
The two possible action spaces are shown hereinafter in Fig.~\ref{fig:action_space}:

\begin{figure}[htbp]
  \begin{center}
    2 phases, action space. \\
    \includegraphics[width=0.8\linewidth,keepaspectratio]{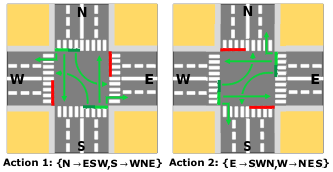}
    \bigskip \\
    4 phases, action space. \\
    \includegraphics[width=0.8\linewidth,keepaspectratio]{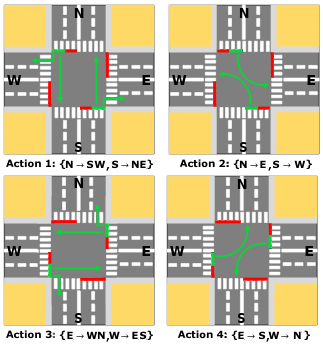}  
    \caption{Action spaces.}
    \label{fig:action_space}
  \end{center}
\end{figure}

\subsection{State representation}

Discrete traffic state encoding (DTSE) \cite{gao2017adaptative} is a microscopic, image-like state representation of the intersection for TSC. Here, we adapt DTSE to the partially observed environment with detection only on CVs, and we propose the partial DTSE.

In partial DTSE, the state representation is an image-like construction of stacked matrices for multiple levels of microscopic, individual information provided by CVs and traffic signals at the intersection. Here, all incoming lanes are discretized, over segments from the stop line up to a detection range less than or equal to the total length of the corresponding approach, into grids with cells of fixed length, set to be slightly larger than the average size of a vehicle with inter-vehicle gap. The cells contain data on individual CVs and traffic signals, and the grids are aggregated over the segments. Three matrices are extracted for three levels of information; i.e. the matrix of CV positions $P$, the matrix of CV speeds $V$, and the matrix of traffic signals over lanes $S$; and stacked into a single 3D image-like state representation: 
\\ 
\[
\text{Partial DTSE} = 
    \begin{bmatrix}
      P = \begin{bmatrix}
      P_0, P_1, P_2, P_3
      \end{bmatrix} \\
      V = \begin{bmatrix}
      V_0, V_1, V_2, V_3
     \end{bmatrix} \\
     S = \begin{bmatrix}
      S_0, S_1, S_2, S_3
     \end{bmatrix}
\end{bmatrix} \]

\begin{figure}[htbp]
  \begin{center}  
    \includegraphics[width=0.8\linewidth,keepaspectratio]{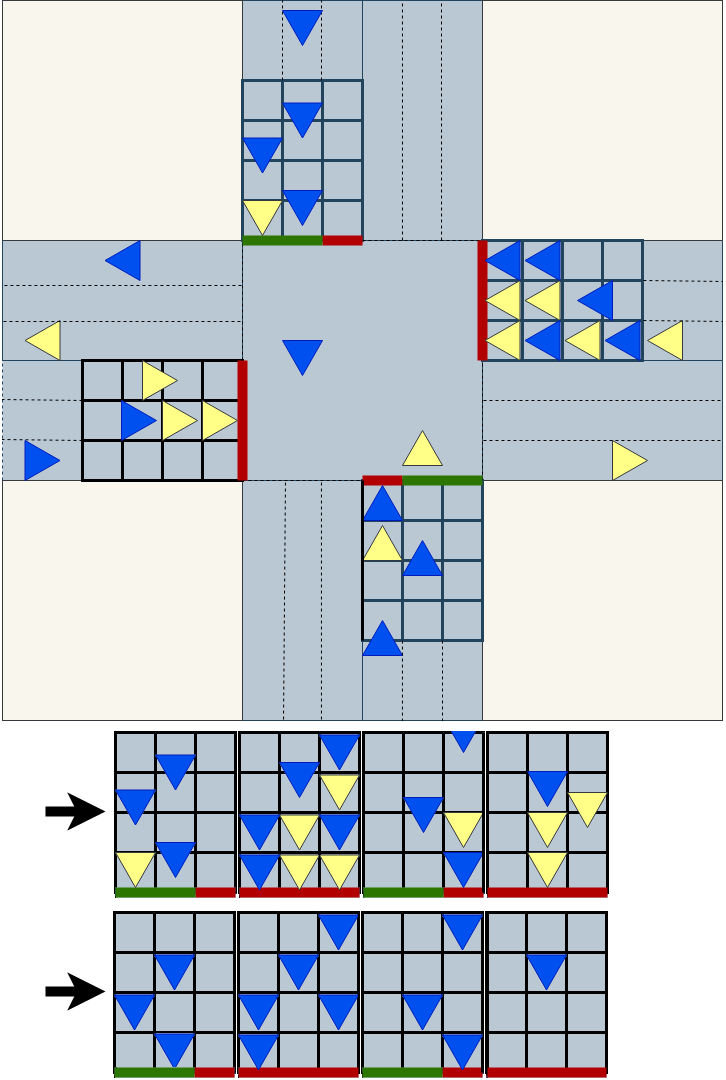}  
  \end{center}
\end{figure}

\begin{table}[!htbp]
\raggedleft
\setlength\tabcolsep{2pt}
P =
\begin{tabular}{|c|c|c|c|c|c|c|c|c|c|c|c|}
\multicolumn{12}{c}{Partial DTSE =} \\
\hline
\multicolumn{3}{|c|}{$P_0$} & \multicolumn{3}{c|}{$P_1$} & \multicolumn{3}{c|}{$P_2$} & \multicolumn{3}{c|}{$P_3$} \\
\hline
\: 0 \: & \: 0 \: & \: 0 \: & \: 0 \: & \: 0 \: & \: 1 \: & \: 0 \: & \: 0 \: & \: 1 \: & \: 0 \: & \: 0 \: & \: 0 \: \\ \hline
\: 0 \: & \: 1 \: & \: 0 \: & \: 0 \: & \: 1 \: & \: 0 \: & \: 0 \: & \: 0 \: & \: 0 \: & \: 0 \: & \: 1 \: & \: 0 \: \\ \hline
\: 1 \: & \: 0 \: & \: 0 \: & \: 1 \: & \: 0 \: & \: 1 \: & \: 0 \: & \: 1 \: & \: 0 \: & \: 0 \: & \: 0 \: & \: 0 \: \\ \hline
\: 0 \: & \: 1 \: & \: 0 \: & \: 1 \: & \: 0 \: & \: 0 \: & \: 0 \: & \: 0 \: & \: 1 \: & \: 0 \: & \: 0 \: & \: 0 \: \\ \hline
\end{tabular}
\\ V = 
\begin{tabular}{|c|c|c|c|c|c|c|c|c|c|c|c|}
\hline
\multicolumn{3}{|c|}{$V_0$} & \multicolumn{3}{c|}{$V_1$} & \multicolumn{3}{c|}{$V_2$} & \multicolumn{3}{c|}{$V_3$} \\
\hline
\: 0 \: & \: 0 \: & \: 0 \: & \: 0 \: & \: 0 \: & \: 0 \: & \: 0 \: & \: 0 \: & 0.4 & \: 0 \: & \: 0 \: & \: 0 \: \\ \hline
\: 0 \: & 0.9 & \: 0 \: & \: 0 \: & 0.2 & \: 0 \: & \: 0 \: & \: 0 \: & \: 0 \: & \: 0 \: & 0.1 & \: 0 \: \\ \hline
 0.8 & \: 0 \: & \: 0 \: & \: 0 \: & \: 0 \: & \: 0 \: & \: 0 \: & 0.7 & \: 0 \: & \: 0 \: & \: 0 \: & \: 0 \: \\ \hline
\: 0 \: & 0.8 & \: 0 \: & \: 0 \: & \: 0 \: & \: 0 \: & \: 0 \: & \: 0 \: & \: 0 \: & \: 0 \: & \: 0 \: & \: 0 \: \\ \hline
\end{tabular}
\\ S = 
\begin{tabular}{|c|c|c|c|c|c|c|c|c|c|c|c|}
\hline
\multicolumn{3}{|c|}{$S_0$} & \multicolumn{3}{c|}{$S_1$} & \multicolumn{3}{c|}{$S_2$} & \multicolumn{3}{c|}{$S_3$} \\
\hline
\: 1 \: & \: 1 \: & \: 0 \: & \: 0 \: & \: 0 \: & \: 0 \: & \: 1 \: & \: 1 \: & \: 0 \: & \: 0 \: & \: 0 \: & \: 0 \: \\ \hline
\: 1 \: & \: 1 \: & \: 0 \: & \: 0 \: & \: 0 \: & \: 0 \: & \: 1 \: & \: 1 \: & \: 0 \: & \: 0 \: & \: 0 \: & \: 0 \: \\ \hline
\: 1 \: & \: 1 \: & \: 0 \: & \: 0 \: & \: 0 \: & \: 0 \: & \: 1 \: & \: 1 \: & \: 0 \: & \: 0 \: & \: 0 \: & \: 0 \: \\ \hline
\: 1 \: & \: 1 \: & \: 0 \: & \: 0 \: & \: 0 \: & \: 0 \: & \: 1 \: & \: 1 \: & \: 0 \: & \: 0 \: & \: 0 \: & \: 0 \: \\ \hline
\end{tabular}
    \captionsetup{justification=centering}
    \captionof{figure}{Partial DTSE.}
    \label{fig:partial_dtse}
\end{table}

The example in Fig.~\ref{fig:partial_dtse} demonstrates partial DTSE applied to a four-way intersection, with CVs in blue and non-CVs in yellow. The matrix of CV positions $P$ is binary encoded, with 1 and 0 being respectively the presence or absence of CVs in the cells. The matrix of CV speeds $V$ encodes the matching speeds, normalized over the speed limits of approaches. The matrix of traffic signals over lanes $S$ is binary encoded, with 1 being a green signal for the lane and 0 a yellow or red signal.

\subsection{Reward function}

The goal of the agent is to minimize the total travel time through the intersection for all commuters, i.e. for both CVs and non-CVs. As such, while the state represents only CVs in a partially observed environment, the reward function considers all vehicles in a fully observed environment. This implies that full training is completed in the simulator, where non-CVs are observable, and performances do not improve after deployment \cite{zhang2020using}. We propose a reward function based on vehicle delay, as delay was considered the most efficient metric for learning by a comparative study \cite{touhbi2017adaptative}. Minimizing delay translates to minimizing the lost travel time $\overline{t} - t_{min}$, with $\overline{t}$ the average travel time of vehicles and $t_{min}$ the lowest possible travel time with speed limit $v_{max}$ \cite{zhang2020using}. At timestep $t$, for a given vehicle $i$ with speed $v_i(t)$, and over the travel distance $t_{min} \cdot v_{max}$ :

\[ t_{min} \cdot v_{max} = \int_{0}^{\overline{t}} v_i(t) \,dt \implies t_{min} = \frac{1}{v_{max}} \int_{0}^{\overline{t}} v_i(t) \,dt \]

\[ \implies \overline{t} - t_{min} = \frac{1}{v_{max}} \int_{0}^{\overline{t}} v_{max} - v_i(t) \,dt \]
\\

Thus, minimizing the total delay is equivalent to minimizing, for each timestep $t$ and vehicle $i$, the individual delay:

\[ d_i(t) = \frac{1}{v_{max}} \cdot ({v_{max}} - v_i(t)) = 1 - \frac{v_i(t)}{v_{max}} \]

We propose here to minimize the total squared delay, the cumulative delay over all incoming vehicles at the intersection, with a power term that prioritizes many short delays over fewer large delays and thereby encourages fairness among vehicles:

\[ tsd(t) = \sum_{i} \left(1 - (\frac{v_i(t)}{v_{max}})^2 \right) \]

The reward function thus maximizes the negative total squared delay. Additionally, the reward is normalized by the maximum total squared delay encountered at that time of training $tsd_{max}(t) = \max (tsd(t), tsd_{max}(t-1))$, and centered in $[0, 1]$ for learning stability \cite{genders2019opensource}, i.e. the TSC reward:

\[ r_t = 1 - \frac{tsd(t)}{tsd_{max}(t)} \]

\subsection{NN architecture}

The convolutional neural network (CNN) \cite{krizhevsky2012imagenet} is a special architecture for image analysis. It is split into a convolutional module, that assembles patterns of increasing complexity in data with space invariant operations, and a fully connected module. A convolutional layer in the convolutional module performs 2D-convolutions over shared weights of filters, i.e. kernels, sliding with a stride along stacked input feature matrices, i.e. channels, to output feature maps. It is defined by a number of channels $C$, a kernel size $K$ and a stride $S$.

Here, partial DTSE is a 3-channels image-like state representation with cells acting as pixels, and we thus use a CNN for the deep $Q$-network, i.e. the convolutional dueling DQN. After experiments, we shaped the CNN to have two convolutional layers (2) $CNN1$ with $(C=16,K=4,S=2)$, (3) $CNN2$ with $(C=32,K=2,S=1)$, and two fully connected layers (4) $FC1$ with $128$ neurons and (5) $FC2$ with $64$ neurons; and all the activations are exponential linear unit (ELU) functions. With (1) the partial DTSE input layer, (6) the dueling stream layer and (7) the $Q$-value aggregation output layer, the convolutional dueling DQN is as in Fig.~\ref{fig:cnn_dueling}:

\begin{figure}[htbp]
  \begin{center}
    \includegraphics[width=1\linewidth,keepaspectratio]{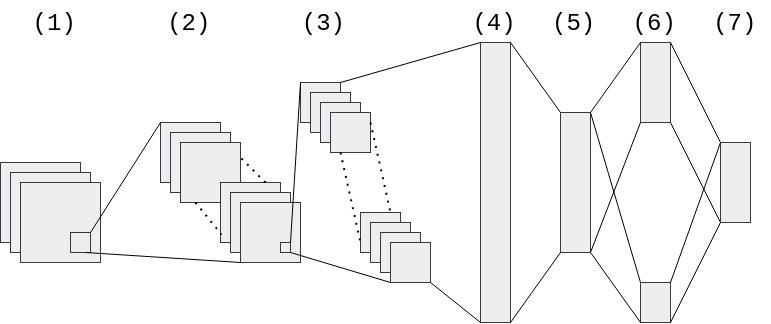}  
    \caption{Convolutional dueling DQN.}
    \label{fig:cnn_dueling}
  \end{center}
\end{figure}

We tuned the model by grid search method over the hyper-parameters, with runs of $2M$ timesteps and episodes of 3600 second simulations. The tuned values are: learning rate $\alpha=\num{1e-4}$ with Adam optimization and Huber loss, discount factor $\gamma=0.99$, minimum epsilon $\epsilon_{min}=0.01$, epsilon decay $\epsilon_{dec}=2$M with exponential decay, replay memory buffer capacity $N=1$M, replay memory buffer initial size $N_{min}=$ $0.1$M and target soft update rate $\tau=\num{1e-3}$ with Polyak update.

\section{Test methodology}
\label{sec:methodology}

\subsection{Simulations}

We trained and tested the proposed model in the SUMO microscopic continuous road traffic simulator \cite{krajzewicz2012recent,kheterpal2018flow}, as the RL environment, on three high-fidelity traffic scenarios, with different road network structures and traffic signal programs:
\begin{enumerate}
  \item Scenario (a): 2 phases, 2x2 incoming lanes;
  \item Scenario (b): 4 phases, 3x3 incoming lanes;
  \item Scenario (c): 4 phases, 4x4 incoming lanes.
\end{enumerate}

In scenarios (a) and (b), one incoming lane is reserved for, respectively permissive and protected, left turns. In scenario (c), two incoming lanes are reserved for protected left turns. For all scenarios, the change, clearance and minimum green intervals are fixed to $T_y = 3$ seconds, $T_r = 2$ seconds and $T_{g,min} = 10$ seconds, and each approach has a length of $300$ meters. As the vehicle size and inter-vehicle gap are $5$ and $2.5$ meters in SUMO, we use cells of $8$ meters in partial DTSE, and a detection range of $160$ meters for up to $20$ CVs per lane. 

Episodes are 3600 second SUMO simulations, with randomly generated traffic demand to create heterogeneous traffic. Each episode is assigned a randomly selected CV penetration rate $p_{cv} \in [0,1]$ and randomly selected insertion traffic flows $q_e \in [100,1000]$ vehicles per hour per entry approach $e$, with traffic demand following a Poisson process with parameter $\lambda_e = q_e^{-1}\cdot 3600$. The vehicle turn ratios are randomly split over the movements, relatively to their number of connections. 

In a learning phase, the DQN agent was first trained for $4M$ timesteps, i.e. 40 hours on an 8-core CPU, in each scenario following an epsilon-greedy policy. In a deployment phase, the DQN agent, with trained neural network weights, was then observed in each scenario following the optimal learned policy.

\subsection{Evaluation process}

In the deployment phase, the performances of the model were evaluated indirectly with a two-step comparative process:
\begin{enumerate}
  \item The efficiency of the model was first assessed in full detection (FD), with $p_{cv} = 1$, by comparing performances with actuated TSC algorithms, Max Pressure and SOTL.
  \item The efficiency of the model was then assessed in partial detection (PD), with $p_{cv} \in [0,1]$, by estimating the loss in performances between FD and PD for levels of $p_{cv}$.
\end{enumerate}

Performances are recorded over 1000 episodes for each of the three scenarios; i.e. (a), (b), (c); and for the four traffic signal controllers; i.e. Max Pressure, SOTL, DQN with FD and DQN with PD. Thus, each algorithm is evaluated over $3 \cdot 1000 = 3000$ hours of diverse traffic situations. Moreover, a random seed was used at deployment, so that controllers experience the same $3000$ hours of exogenous traffic and are comparable by episodes, to evaluate global tendencies in flows.

We measure the performances by the average over an episode of the total delay at the intersection for all vehicles, as episodes are unitary traffic situations and for $3000$ comparison points per algorithm: the episode mean total delay (EMTD).

\subsection{Comparison algorithms}

As there exists no standard algorithm for deep Q-learning TSC with partial detection, the performances of the model are evaluated in full detection against two actuated TSC (ATSC) algorithms: Max Pressure and SOTL. They are adaptive, non-learning TSC algorithms, responding to traffic flows in real time by measuring requests for green signals over the competing phases, according to fixed rules \cite{wei2020survey}. They rely on full, macroscopic traffic detection and are already deployed at many real intersections with inductive loops under the roads.
\\

\subsubsection{Max Pressure} \cite{varaiya2013max, genders2019opensource} is an acyclic ATSC algorithm which minimizes the pressure of phases at an intersection. The pressure of a phase $p$ is defined as the difference between the total number of vehicles in the set of all incoming lanes with authorized movements for that phase $L_{p,inc}$ and the total number of vehicles in the set of all corresponding outgoing lanes $L_{p,out}$. After each minimum green interval of time $T_{g,min}$, the controller selects the phase $p \in P$ with maximum pressure to be relieved in the set of all possible phases. While Max Pressure is efficient and simple to implement, it involves detection on vehicles in both incoming and outgoing lanes and is thus costly to deploy. 
\\

\begin{algorithm}[H]
\caption*{Max Pressure algorithm}
\begin{algorithmic}
    \STATE \textbf{procedure} MaxPressure($T_g$,$T_{g,min}$):
    \IF{$T_g \ge T_{g,min}$}
        \STATE Set next phase to maximum pressure phase
        \STATE $p = arg\max(\{pressure(p) \textbf{ for } p \in P\})$
        \STATE with $pressure(p) = \sum_{l \in L_{p,inc}} |V_l| - \sum_{l \in L_{p,out}} |V_l|$;
    \ENDIF
\end{algorithmic}
\end{algorithm}

\subsubsection{SOTL} \cite{gershenson2004self, genders2019opensource} (Self-organizing traffic lights) is a cyclic ATSC algorithm which dynamically sets the green interval duration $T_g$ in the current phase $p$. A time integral $\chi$ of the total number of vehicles in the set of all incoming lanes with prohibited movements for that phase $L_{inc} - L_{p,inc}$ and in a distance $\psi$ from the stop line is accumulated, and the current phase is maintained until the accumulated time integral reaches a fixed threshold $\mu$. Additionally, small vehicle platoons of size $\eta$, the total number of vehicles in the set of all incoming lanes with authorized movements for that phase $L_{p,inc}$ and in a distance $\omega \le \psi$ from the stop line, are kept together and prevent phase changes for sizes less than a fixed threshold $\nu$. 

Here, we set the constants to values commonly used in the literature: $\mu = 50$, $\nu = 3$, and $\psi = 80$, $\omega = 25$ meters \cite{placzek2014self}. \\

\begin{algorithm}[H]
\caption*{SOTL algorithm}
\begin{algorithmic}
    \STATE \textbf{procedure} SOTL($T_g$,$T_{g,min}$,$\mu$,$\nu$,$\psi$,$\omega$):
    \STATE Accumulate time integral $\chi = \chi + \sum_{l \in L_{inc} - L_{p,inc}} |V_{l,\psi}|$;
    \IF{$T_g \ge T_{g,min} \textbf{ and } \chi > \mu$}
        \STATE Set vehicle platoon size $\eta = \sum_{l \in L_{p,inc}} |V_{l,\omega}|$;
        \IF{$\eta \text{ null  }\textbf{ or } \eta > \nu$}
            \STATE Set $\chi = 0$;
            \STATE Set next phase to next phase in cycle $p \in P$;
        \ENDIF
    \ENDIF 
\end{algorithmic}
\end{algorithm}

\section{Results and discussion}
\label{sec:results}

\begin{figure}[htbp]
  \begin{center}
    \includegraphics[width=0.9\linewidth,keepaspectratio]{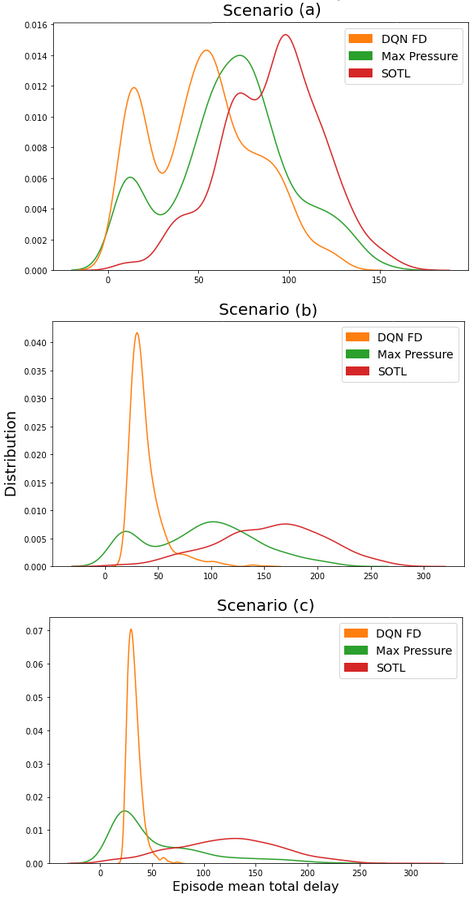}  
    \caption{EMTD probability distributions for DQN with FD, Max Pressure and SOTL; by value, and by scenario.}
    \label{fig:fd_1}
  \end{center}
\end{figure}

\begin{figure}[htbp]
  \begin{center}
    \includegraphics[width=0.9\linewidth,keepaspectratio]{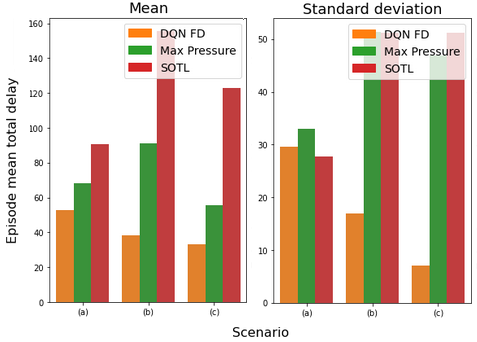}  
    \caption{EMTD mean and standard deviation for DQN with FD, Max Pressure and SOTL; by scenario.}
    \label{fig:fd_2}
  \end{center}
\end{figure}

\begin{figure}[htbp]
  \begin{center}
    \includegraphics[width=0.9\linewidth,keepaspectratio]{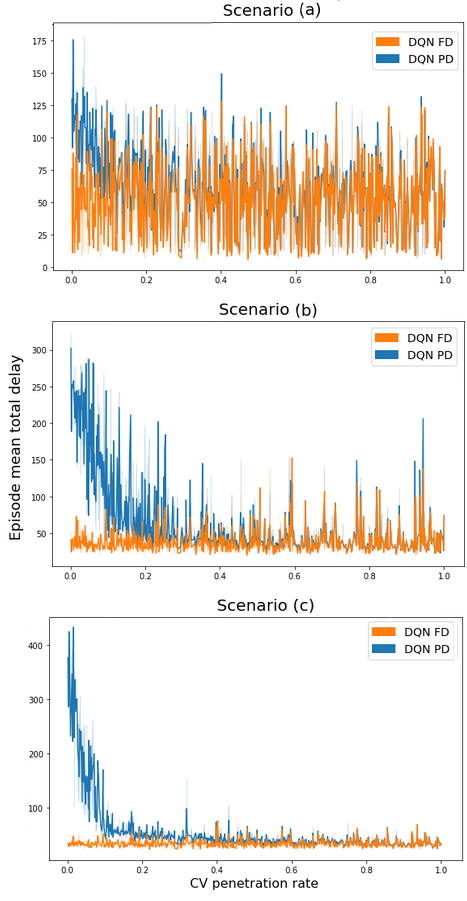}  
    \caption{EMTD for DQN with FD and DQN with PD; by CV penetration rate, and by scenario.}
    \label{fig:pd_1}
  \end{center}
\end{figure}

\begin{figure}[htbp]
  \begin{center}
    \includegraphics[width=1\linewidth,keepaspectratio]{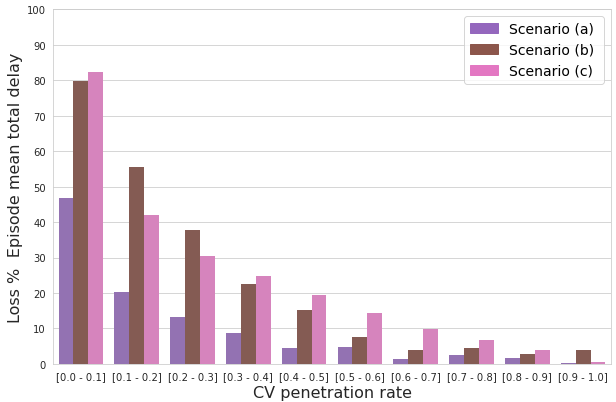}  
    \caption{EMTD loss (\%) between DQN with FD and DQN with PD for all scenarios; by range of CV penetration rate.}
    \label{fig:pd_2}
  \end{center}
\end{figure}

\subsection{Analysis}

We evaluate the performances of the model by EMTD over $3000$ hours of simulation, in a two-step comparative analysis.

(1) We first assess the performances in full detection, by comparing DQN with FD to Max Pressure and SOTL. Fig.~\ref{fig:fd_1} compares the probability distribution functions by values and scenarios. Fig.~\ref{fig:fd_2} compares the means and standard deviations by scenarios. (2) We then assess the performances in partial detection, by comparing DQN with PD to DQN with FD. 
Fig.~\ref{fig:pd_1} compares values between FD and PD by CV penetration rates and scenarios. Fig.~\ref{fig:pd_2} estimates the average loss between FD and PD by scenarios and ranges of CV penetration rates.

\subsubsection{Full detection}
From Fig.~\ref{fig:fd_1} and Fig.~\ref{fig:fd_2}, DQN with FD performs better than the two ATSC algorithms in all scenarios. 

In scenario (a) with a 2-phases program, a small improvement is visible in the probability distribution curve. The mean is slightly lower, with a standard deviation similar to those of Max Pressure and SOTL. We assume that this scenario is simple and does not allow for a relevant margin of progress and further optimization past actuated methods, and that the small size of the road network has led to over-saturation in many episodes with high demands. Nonetheless, performances are at least equivalent to ATSC, and this with cost-efficiency.

In scenarios (b) and (c) with 4-phases programs, DQN with FD outperforms significantly Max Pressure and SOTL. While both means are strongly lower, the most notable results here are the differences in standard deviations, which are low for DQN with FD and high for the ATSC algorithms. This is confirmed in the probability distribution curves, which form dense spikes concentrated around lower values for DQN with FD, while being spread out and flattened over wider ranges of values for both Max Pressure and SOTL. This implies that DQN with FD is more robust towards a great diversity of traffic situations than the actuated methods, for complex intersections with varying demands and 4-phases programs.\\

\subsubsection{Partial detection}
From Fig.~\ref{fig:pd_1} and Fig.\ref{fig:pd_2}, DQN with PD converges towards FD, as the CV penetration rate increases. 

The performances in PD diverge strongly for CV penetration rates less than 10\%. They approach asymptotically the performances for FD for CV penetration rates from 10\% to 40\%, and the performances are quasi identical with CV penetration rates higher than 40\%, from DQN with PD to DQN with FD.

We estimate the average loss of DQN with PD, i.e. the average performance gain of FD over PD, in a range of CV penetration rates, in percentage. For very low CV penetration rates less than 10\%, the performance loss is very high, 50\% to 80\%. It then decreases markedly in the following ranges of CV penetration rates from 10\% to 40\%, being approximately in 20\% to 40\%. Finally, it stabilizes under 20\% starting from CV penetration rates in 40\% to 50\%, and then decreases with increasing CV penetration rates until reaching DQN with FD.

\subsection{Observation in SUMO}
We observed the trained, deployed DQN agent controlling traffic signals in SUMO with the optimal learned policy. It effectively learned that phase changes induce wasted time due to the intermediate transitions through change and clearance intervals, and thus maintains the ongoing phases for sufficiently long green intervals to avoid flickering, and to minimize the accumulated waiting time and the number of stops in a queue. Yet, the agent also learned to promote fairness among phases, and lanes with lower demand are not blocked in favor of minimum travel times in lanes with higher demand. Moreover, it learned to instantly adapt to changing flows and to balance phase changes adequately to maximize speed for all incoming vehicles, thus minimizing total delay. This learned policy performs efficiently for CV penetration rates as low as 20\%, which appears as the lower bound to infer traffic flow parameters from the CVs in a partially observed environment. 

Else, for CV penetration rates under 20\%, the DQN agent performs poorly, and due to the learned cost of phase change and the few detected vehicles, traffic signals are maintained for very long times, creating jams in all incoming approaches and locking the intersection. 
This issue can be limited in practice by defining a maximum green interval $T_{g,max}$, after which a phase change is enforced in a cycle, thus the controller acting as a pretimed fixed program for very low CV penetration rates.

\subsection{Summary}
We summarize the performances and draw two main results:
\begin{enumerate}
  \item From the comparative analysis in full detection and the observations in SUMO, the proposed DQN model substantially improves the performances of TSC compared to deployed actuated methods Max Pressure and SOTL, achieving both fairness between vehicles and global efficiency at the intersection. Moreover, the DQN model is more robust and adaptive to a high diversity of traffic situations, for more complex intersection configurations with varying traffic demands and for 4-phases programs.
  \item From the comparative analysis in partial detection and the observations in SUMO, the proposed DQN model functions for CV penetration rates from 20\% upwards. Thus, we propose two performance thresholds: (1) the acceptability threshold, whereupon the DQN model becomes advantageous to deploy in PD, at CV penetrations rates $p_{cv} \geq 20\%$, and (2) the optimality threshold, whereupon the DQN model in PD attains high efficiency similar to FD, at CV penetrations rates $p_{cv} \geq 40\%$.
\end{enumerate}

\section{Conclusion and perspectives}
\label{sec:conslusion}

\subsection{Conclusion}
In this paper, we presented a novel model for deep Q-learning traffic signal control at single intersections with partial detection over connected vehicles. We introduced a new state representation for partially observable environments, partial DTSE, and a new reward function for TSC, total squared delay. Additionally, we provided tuned values for the convolutional dueling DQN architecture and hyper-parameters.

We evaluated the model performances against two existing actuated controllers, Max Pressure and SOTL, in a two-step comparative analysis by episode mean total delay. As a result, we concluded the model to be more efficient than the ATSC algorithms for 4-phases programs in full detection, and estimated partial detection performance thresholds for CV penetration rates: acceptability at $20\%$ and optimality at $40\%$.

\subsection{Perspectives}
This work can be pursued in many ways: 
(1) We assumed information on CVs to be exact despite imprecise infrastructures, and probabilistic estimation methods could be used to compensate for the uncertainty in measurements. 
(2) We trained a new agent for each scenario, and we think that partial DTSE could be generalized across multiple intersection sizes with zero-padding. 
(3) We proposed a reward function over all vehicles that learns entirely in the simulator, while one based only on CVs could continue to optimize policies after deployment at a real intersection.
(4) We suggested to limit degraded performances for low CV penetration rates with a maximum green interval, but a more efficient solution appears as necessary. 
(5) We provided a model for single agent TSC, and we believe that it could be extended to decentralized multi agent RL for communicating grids of coordinated traffic lights.
(6) We tested the model only on synthetic edited data, and it could be further validated on real world simulated networks, e.g. with the Luxembourg SUMO traffic (LuST) scenario \cite{Codeca2015luxembourg}.

Furthermore, other network configurations; e.g. 3-way and 5-way intersections; or road parameters; e.g. prioritized vehicles, pedestrian crossings, public transports; could be explored.

From these perspectives and the results discussed before, we find the proposed model to be both effective and promising.

\printbibliography

\end{document}